\documentclass[bst/sn-mathphys,Numbered]{sn-jnl}
\usepackage{graphicx}%
\usepackage{multirow}%
\usepackage{amsmath,amssymb,amsfonts}%
\usepackage{amsthm}%
\usepackage{mathrsfs}%
\usepackage[title]{appendix}%
\usepackage{xcolor}%
\usepackage{textcomp}%
\usepackage{manyfoot}%
\usepackage{booktabs}%
\usepackage{algpseudocode}%
\usepackage{listings}%

\usepackage{amsmath,amsfonts}
\usepackage{array}
\usepackage{textcomp}
\usepackage{stfloats}
\usepackage{url}
\usepackage{verbatim}
\usepackage{graphicx}
\usepackage{cite}
\usepackage[ruled,vlined]{algorithm2e}
\SetKwInput{KwInput}{Input}
\SetKwInput{KwOutput}{Output}

\usepackage{booktabs}
\usepackage{multirow}
\usepackage{subfigure}
\usepackage{epsfig}
\usepackage{amssymb}
\usepackage{bm, eucal}
\usepackage{natbib}
\usepackage{xcolor}
\usepackage[normalem]{ulem}

\DeclareMathOperator*{\argmin}{arg\,min}

\newcommand{\etal}{\textit{et al.}}

\theoremstyle{thmstyleone}%

\theoremstyle{thmstyletwo}%

\theoremstyle{thmstylethree}%

\begin{document}

\title[Article Title]{Kernel Adversarial Learning for Real-world Image Super-resolution}

\author[1]{\fnm{Hu} \sur{Wang}}\equalcont{}
\author[1]{\fnm{Congbo} \sur{Ma}}\equalcont{Contributed equally to this work.}
\author[2]{\fnm{Jianpeng} \sur{Zhang}}
\author[1]{\fnm{Wei Emma} \sur{Zhang}}
\author[3]{\fnm{Gustavo} \sur{Carneiro}}

\affil[1]{\orgaddress{The University of Adelaide, 5005 Adelaide, Australia}}
\affil[2]{\orgaddress{Alibaba DAMO Academy, Hangzhou, China}}
\affil[3]{\orgaddress{Centre for Vision, Speech and Signal Processing, University of Surrey, United Kingdom}}

%%==================================%%
%% sample for unstructured abstract %%
%%==================================%%

\abstract{
Current deep image super-resolution (SR) approaches aim to restore high-resolution images from down-sampled images or by assuming degradation from simple Gaussian kernels and additive noises. However, these techniques only assume crude approximations of the real-world image degradation process, which should involve complex kernels and noise patterns that are difficult to model using simple assumptions. In this paper, we propose a more realistic process to synthesise low-resolution images for real-world image SR by introducing a new Kernel Adversarial Learning Super-resolution (KASR) framework. In the proposed framework, degradation kernels and noises are adaptively modelled rather than explicitly specified. Moreover, we also propose a high-frequency selective objective and an iterative supervision process to further boost the model SR reconstruction accuracy. Extensive experiments validate the effectiveness of the proposed framework on real-world datasets.
}

\keywords{Image super-resolution, adversarial learning, degradation kernels, noises}

\maketitle

%% main text
\section{Introduction}

In the image super-resolution (SR) task, low-resolution images will be inserted into an SR model and the corresponding high-resolution images are expected to be restored. SR techniques can be applied to a wide range of real-world applications, e.g. high-definition (HD) television displays, zooming processes on cell-phones or drones~\citep{wang2021fully}. Similar to many computer vision applications, SR methods are currently mostly based on deep learning approaches. Dong \etal~\citep{dong2015image} proposed the first deep learning method using a three-layer convolutional neural network for the SR task. Since then, deep learning SR research has flourished, with the proposal of many powerful models~\citep{kim2016accurate,ledig2017photo,lim2017enhanced,zhang2018image,wang2018esrgan}.

According to~\citep{wang2020deep}, the SR process can be explained with the simple formulation below. The low-resolution image can be obtained by a process that combines a degradation kernel with additive noise, formulated as follows:
\begin{equation}\label{eqn:degradation}
	I_{LR}=(I_{HR} \ast k)\downarrow_s + \delta,
\end{equation}
where a low-resolution (LR) image $I_{LR}$ is formed from a high-resolution image $I_{HR}$ by blurring it with kernel $k$ (using the convolution operator $\ast$), then down-sampling it with scale $s$ using the operator $\downarrow_s$, and adding noise $\delta$.

Although useful for training SR models, the degradation process in~\eqref{eqn:degradation} is a crude approximation of real-world image degradation. Nevertheless, naive image down-sampling and Gaussian blur kernels combined with additive Poisson or Gaussian noises are still adopted to produce low-resolution images for the construction of SR datasets because of their simplicity. A more realistic degradation process is more difficult to be modelled because real-world sRGB noise is formed by an image signal processing (ISP) pipeline with a series of non-linear operations. In 2000, Tian \etal~\citep{tian2001analysis} analysed the temporal noise caused by CMOS image sensors. Similarly, in 2006, Liu \etal~\citep{liu2006noise,liu2007automatic} studied the relationship between changes of noise level and brightness. They also attempted to infer the noise level and function from an image with Bayesian MAP inference. However, the noise types they tried to model turned out to be extremely non-additive, where the noise types and levels varied across different CCD digital cameras.

Many existing SR papers~\citep{dong2015image,kim2016deeply,shi2016real,ledig2017photo,lim2017enhanced,zhang2018image} assume that low-resolution images are formed by down-sampling high-resolution images with naive bicubic interpolation. This strong assumption results in an inconsistency between the synthetically degraded training images using Eq.\eqref{eqn:degradation} and the naturally degraded real-world images. In real-world image super-resolution scenarios, arbitrary highly non-additive noises may be generated from motion blur, defocus, image compression, or simply because of particular characteristics associated with CMOS image sensors. Such training and testing image inconsistencies can cause a poor generalisation of the SR model trained on bicubic down-sampled (or Gaussian-kernel degraded) LR images.

Some existing super-resolution models~\citep{bell2019blind,shocher2018zero} attempt to alleviate this problem by estimating the real-world kernel from a few-shot learning perspective. Another typical solution~\citep{ji2020real} is to maintain a finite set of estimated degradation kernels and additive noises to accommodate the SR process. However, none of the models above can be applied directly in the inference phase because they require to be trained with testing images. Also, they still need to estimate linear kernels that may not reliably represent a complex degradation process in the real world.

\begin{figure}[]
\centering
\scalebox{1.0}{
\centerline{\includegraphics[width=1\textwidth]{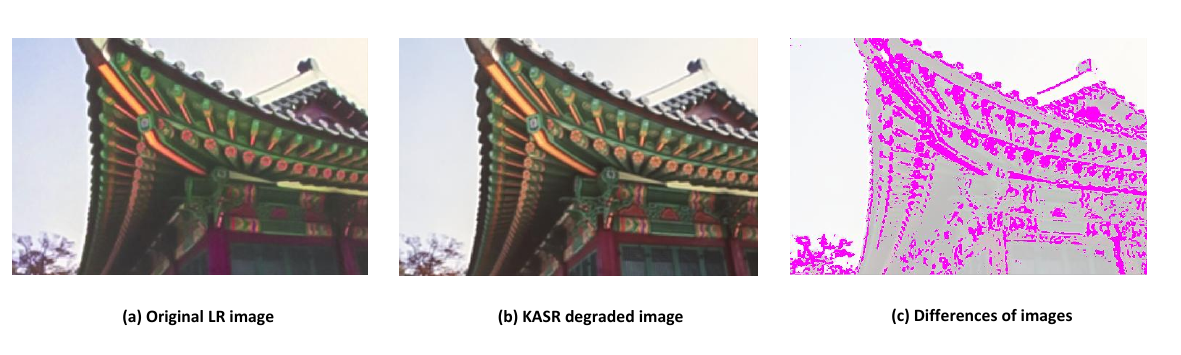}}
}
% \vspace{-2mm}
\caption{Intuitive visualisation of the original LR image and degraded LR image produced by the KASR model. The image in (c) highlights (in pink) the differences between the original low-resolution image in (a) and the LR image down-sampled by the proposed KASR model in (b). Notice that the differences are mainly due to colour changes combined with some texture distortions.
}
\label{fig:gen-img}
% \vspace{-2mm}
\end{figure}

In this paper, we introduce a generic framework to better represent the real-world image super-resolution process, named \underline{K}ernel \underline{A}dversarial Learning \underline{S}uper-\underline{R}esolution (KASR). KASR can adversarially generate degradation kernels and inject noises, but without requiring prior knowledge of the kernels or noises. It is designed to allow a seamless integration with many mainstream SR models. The LR images generated by KASR and the differences with the original LR ones are shown in Figure~\ref{fig:gen-img}. Also, a high-frequency selective loss function is proposed to further force the model to attend to high-frequency areas within SR images. Furthermore, we introduce an optimisation process named Iterative Supervision to gradually refine the SR images. Experimental results demonstrate that, when compared to multiple strong baselines on real-world datasets, mainstream SR models equipped with KASR are able to generate high-quality SR images in real-world scenarios. Our contributions are summarised as follows:

\begin{itemize}
\item We propose the Kernel Adversarial Learning Super-Resolution (KASR) framework to deal with the real-world degradation kernel estimation issue from an adversarial training perspective. The proposed framework is able to model the image degradation process implicitly (i.e., without explicitly specifying a finite set of kernels and noises~\citep{ji2020real}) by exploring an end-to-end training of neural networks.
\item Given the need of SR models to focus on high-frequency areas of the image, we design a high-frequency selective objective function to assign more attention to high-frequency regions of the images during the minimisation of the super-resolution reconstruction error.
\item To effectively utilise existing supervision signals, we introduce the Iterative Supervision optimisation process to gradually refine the SR image.
\end{itemize}

We show extensive experiments which demonstrate that  mainstream SR models equipped with KASR achieve superior performance over competitive methods across two real-world datasets.

\section{Related Work}
\subsection{Real-world Image Super-resolution}

Low-resolution images can be viewed as degraded high-resolution images after blurring, down-sampling and noise interference. An important premise of deep learning, and machine learning in general, is that the distribution from which test images are drawn is the same as the distribution used to generate the training data. When the naive degradation process in the training phase is inconsistent with the degradation process in real-world images, the SR result will not be accurate. Therefore, the simulation of a real-world complex degradation process is crucial.

Gaussian kernel is the most widely adopted blur kernel~\citep{yang2014single,dong2012nonlocally}, but it differs from real-world blur kernels significantly. Therefore, some papers attempted to simulate the degradation process in a more complicated manner by considering a variety of degradation kernels and noise levels. Zhang \etal~\citep{zhang2020deep} examine the performance of SR models under a set of Gaussian blur kernels and receives different results under different kernel assumptions. Another typical idea is to estimate a finite set of down-sampling kernels/noises and inject them into training images as~\citep{ji2020real}. However, such finite set of kernels and noises still represent a strong assumption.

Approaching the problem from a few-shot learning perspective, Shocher \etal~\citep{shocher2018zero} proposed a kernel estimation approach by further down-sampling the given low-resolution image and learning a super-resolution function from these low-high resolution image pairs. However, this method is not practical to be applied directly in the inference phase (e.g. on cell-phone zooming) since training is required on each test image, which is time-consuming. A similar idea has been adopted by KernelGAN~\citep{bell2019blind}. Nevertheless, in the  models above, the unrealistic assumption of Gaussian blur kernels still exists in model training.

\subsection{Adversarial Training}

As shown in~\citep{goodfellow2014explaining}, deep neural network models are very sensitive to small perturbations of the inputs. In particular, tiny pixel-wise distortions can lead deep neural networks to produce entirely different results. This issue can be mitigated with adversarial training that trains models to be robust against attacks formed by dynamically and adaptively adding perturbations to the training samples.

In real-world SR scenarios, models usually suffer from poor generalisation performance due to unknown degradation kernels and noises in the inference phase. Furthermore, non-additive noises can be injected by different CMOS image sensors. Although GANs~\citep{wang2018esrgan,ren2020real} and the idea of adversarial attacks~\citep{choi2019evaluating,choi2020adversarially,yue2021robust} have been introduced in SR to serve as objective functions for the generation of more photo-realistic images, they were used in a totally different context (from the perspective of model security to defence adversarial attacks) from what we are exploring. Existing work~\citep{kim2020dual} adopt down-scaling process of low-resolution image to tackle real-world super-resolution task. However, it forms an cycle reconstruction learning of down-scaling network and SR network which is totally different from our proposed model that adversarially injecting noises and estimate unknown kernels. 
In our paper, we adopt adversarial training to model unknown kernels and non-additive noises for real-world image super-resolution. To the best of our knowledge, for the purpose of implicit simulation of the degradation process, we are the first to introduce adversarial training into the SR task for fulfil this purpose.

\section{The Proposed Method}

\begin{figure}[]
\centering
\scalebox{1.0}{
\centerline{\includegraphics[width=1\textwidth]{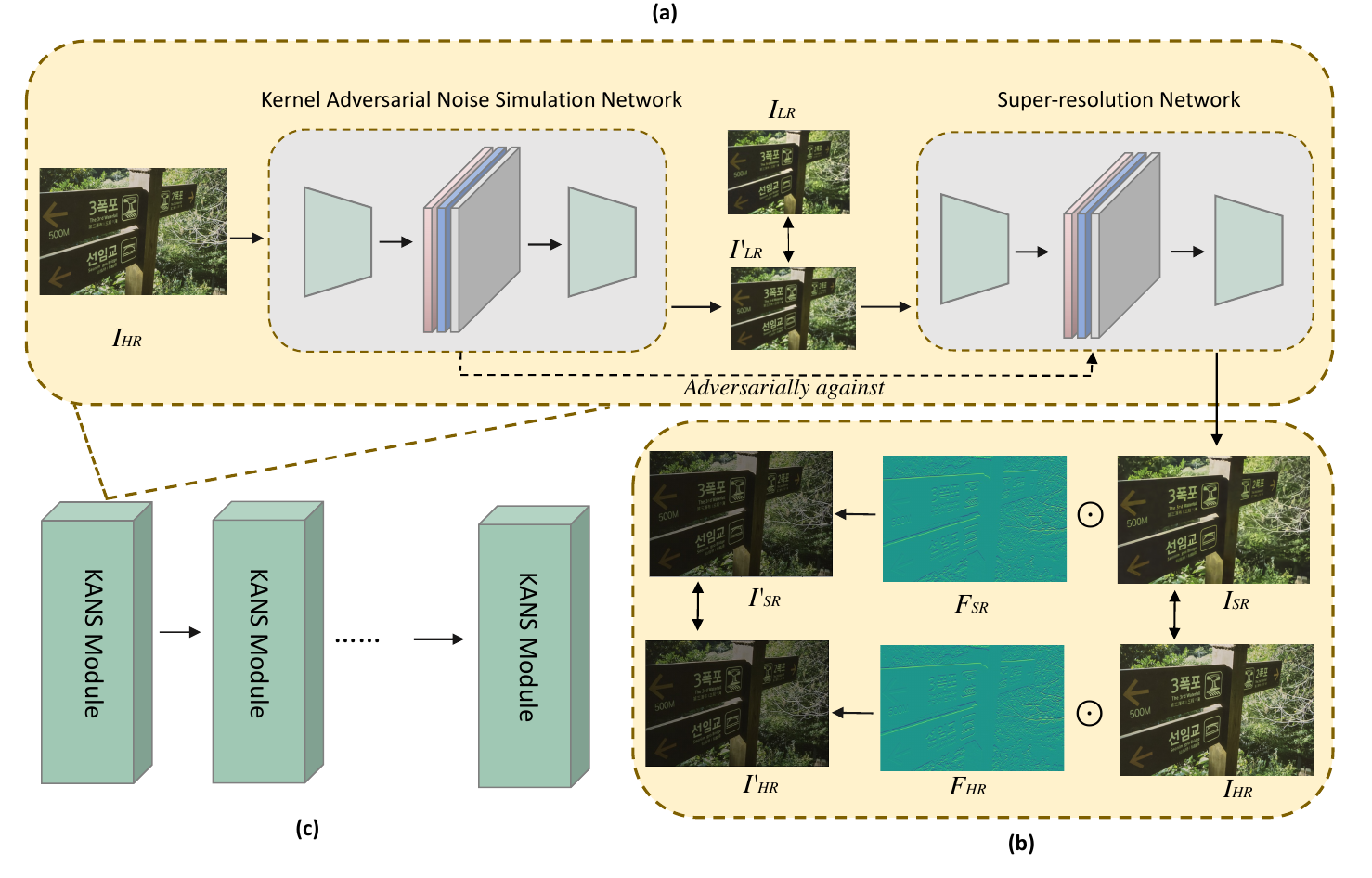}}
}
% \vspace{-2mm}
\caption{The overall framework of the proposed KASR framework. The framework consists of three parts: (a) the Kernel Adversarial Noise Simulation (KANS) to adaptively simulate the image degradation process; 
(b) the High-frequency Selective Objective to force the model to focus on high-frequency regions within images due to the higher importance placed in the image regions containing high-frequency for the SR task; and 
(c) the stacking of multiple KANS modules, where the Iterative Supervision (IS) can leverage the supervision signals to accurately refine the SR image reconstruction.
}
\label{fig:framework}
%\vspace{-2mm}
\end{figure}

\subsection{Overall Framework}

The proposed KASR framework is divided into three parts: the Kernel Adversarial Noise Simulation (KANS) to implicitly and dynamically simulate the image degradation process for robust SR model training; the High-frequency Selective Objective (HFSO) to constrain the model to focus on high-frequency regions of the images; and the Iterative Supervision (IS) to leverage the provided supervision signals in a better manner by repeatedly stacking KANS modules for SR reconstruction refinement. These parts are designed to complement each other and to be combined seamlessly with mainstream SR models.

\noindent\textbf{Kernel Adversarial Noise Simulation.} As shown in Figure \ref{fig:framework}, a high-resolution image $I_{HR}$ is fed into the KANS network to generate a low-resolution image $I'_{LR}$ perturbed with non-additive noises. We constrain the $I'_{LR}$ to be visually similar to the original low-resolution image $I_{LR}$, but the injected perturbations will try to interfere with the downstream super-resolution image restoration process by maximising the super-resolution loss.

\noindent\textbf{High-frequency Selective Objective.} After the super-resolution image is generated, the HFSO constrains the model to focus on the high-frequency regions of the images since an accurate reconstruction in those regions is crucial for a reliable SR model. In HFSO, we extract high-frequency maps from the HR and SR images and  normalise them to be in the range $[0,1]$ to form the high-frequency attention masks $F_{SR}$ and $F_{HR}$. Next, we apply the Hadamard product between the high-frequency attention masks and the HR/SR images to get the high-frequency filtered images. Moreover, the IS iteratively refines the results.

\noindent\textbf{Iterative Supervision.} In standard SR models, high-resolution images are directly generated from low-resolution images. Different from that, we propose to perform iterative optimisation by repeating the degradation and the super-resolution processes. The ground-truth high-resolution images will be used to train each KANS module. The gradients are able to flow all the way back through the chain by providing supervision to the iteratively generated images. Therefore, supervision signals can be more effectively used to better supervise the super-resolution process.

\subsection{Kernel Adversarial Noise Simulation}

In this section, we propose the Kernel Adversarial Noise Simulation (KANS) module. Adversarial training techniques were initially proposed to defend against adversarial attacks~\citep{goodfellow2014explaining,tramer2017ensemble}. However, in the context of real-world SR, we adopt adversarial training to actively generate adaptive blur kernels and non-trivial simulated noises to be injected into the low-resolution image with the goals of reconstructing well the low resolution image and adversarially reconstructing the super-resolution image. This problem is formulated as an optimisation problem to search for the suitable blur kernel $k$ and noise $\delta$ subject to get the optimal parameters $\theta_\eta^*$ for super-resolution. This optimisation is a minimisation maximisation problem that adversarially reconstructs the super-resolution image, as follows:

\begin{equation}\label{eqn:optim}
 \begin{split}
 \min_{k, \delta} \;\;\; &  \| ((I_{HR}\ast k)\!\downarrow_{\bf{s}} + \delta) - I_{LR} \|_p - \\  
 & \| \eta((I_{HR}\ast k)\!\downarrow_{\bf{s}} + \delta; \theta_\eta^*) - I_{HR} \|_p \\
 s.t. \;\;\; & \theta_\eta^* = \argmin \| \eta((I_{HR}\ast k)\!\downarrow_{\bf{s}} + \delta; \theta_\eta) - I_{HR} \|_p,
 \end{split}
\end{equation}
where $s$ is the scale, $(I_{HR}\ast k)\!\downarrow_{\bf{s}}$ is the degradation process, $\eta(\cdot)$ is the super-resolution function parameterised by $\theta_\eta$, and $||\cdot||_p$ denotes the p-norm operator. In super-resolution, the direct optimisation of Eq.\ref{eqn:optim} is not trivial because we need to define an appropriate domain for the kernels and noises; it can be computationally expensive as well. Alternatively, we propose to solve this optimisation by adopting neural networks to directly generate the degradation kernel and the noise to down-sampled images, such that the noise can be generated dynamically with little computational increment. Hence, the optimisation from Eq.\ref{eqn:optim} can be re-formulated as the following two-step optimisation:

\begin{equation}\label{eqn:KANS}
\scalebox{0.9}{$
\begin{split}
\text{Step 1: } & \min_{\theta_\phi} 
\| (\phi(I_{HR}; \theta_\phi) - I_{LR} \|_p - \| \eta(\phi(I_{HR}; \theta_\phi); \theta_\eta) - I_{HR} \|_p, \\
\text{Step 2: } & \min_{\theta_\eta} 
\| \eta(\phi(I_{HR}; \theta_\phi); \theta_\eta) - I_{HR} \|_p,
\end{split}
$}
\end{equation}
where the $\phi(\cdot)$ is the degradation neural network function with parameter $\theta_\phi$.
By doing so, complex blur kernels and highly non-additive noises can be injected into the generated low-resolution image. More specifically, as mentioned above, a data-driven approach is adopted to solve the non-additive blur kernel and noise injection problem by minimising the reconstruction error over the generated and real low-resolution image pairs but maximising the super-resolution image restoration error.

\subsection{Objective Functions}

In the super-resolution task, high-frequency (HF) areas of an image ought to be explored more than low-frequency (LF) areas to improve the reconstruction quality. Naive pixel-wise losses (e.g., MSE objective) assign the same weight to each pixel, independently if it is part of a low or high-frequency region, and average the errors that come from these pixels. Inevitably, this strategy will result in blurred SR image reconstruction because images tend to have a disproportionally larger amount of LF regions than of HF regions, so these LF regions will be better reconstructed than the HF areas, resulting in such blurred reconstruction. Thus, pixels should be assigned with different weights depending if they lie in an LF or HF region of the image for SR tasks. Due to the imbalanced distribution of HF and LF regions, pixels located in HF regions require more attention. Therefore, we adopt the HFSO to attend more to the HF regions within an image. Formally, the HFSO is represented as:

\begin{equation}\label{eq:l_hfso}
  L_{HFSO}(\theta_{\eta}) = \| \psi(f(I_{HR})) \odot I_{HR} - \psi(f(I_{SR})) \odot I_{SR} \|_p,
\end{equation}
where the super-resolution image $I_{SR} = \eta(\phi(I_{HR};\theta_{\phi});\theta_{\eta})$ has the same dimensions as $I_{HR}$, 
$f(\cdot)$ is the high-frequency filter, $\psi(\cdot)$ denotes the normalisation operator, and $\odot$ represents the element-wise Hadamard product. Similar to Equation~\eqref{eqn:KANS}, $p$-norm is performed in the Equation~\eqref{eq:l_hfso}.

Besides the aforementioned objectives of KANS and HFSO, the pixel-wise super-resolution reconstruction objective has been adopted:
\begin{equation}\label{}
  L_{REC}(\theta_{\eta}) = \| I_{SR} - I_{HR} \|_p,
\end{equation}
where $I_{SR}$ is defined in Equation~\eqref{eq:l_hfso}.

Therefore, the total objective function for the super-resolution network is:
\begin{equation}\label{}
  L_{SR}(\theta_{\eta}) = L_{REC}(\theta_{\eta}) + \omega L_{HFSO}(\theta_{\eta}),
  \label{eq:L_SR}
\end{equation}
where $\omega$ is the trade-off factor between this two objectives. From the adversarial training perspective, besides the Equation \ref{eqn:KANS}, a discriminative objective $L_{d}$ is also adopted to train a discriminator to classify the original and generated LR images into real or fake. This is designed to further ensure the visual quality of the generated images. So the overall objective for KANS can be presented as:

\begin{equation}\label{}
  L_{KANS}(\theta_{\phi}) =  \| I'_{LR} - I_{LR} \|_p - \beta\| I_{SR} - I_{HR} \|_p + \gamma L_{d},
  \label{eq:L_KANS}
\end{equation}
where $I'_{LR} = \phi(I_{HR}; \theta_{\phi})$, $\beta$ and $\gamma$ are weight factors to balance the three terms. The training process of KASR can be found in Algorithm \ref{alg}. 

\begin{algorithm}[]
	\DontPrintSemicolon
	\SetAlgoLined
	
	\KwInput{Low-resolution image $I_{LR}$,  degradation neural network parameter $\theta_\phi$, super-resolution network parameter $\theta_\eta$, number of KANS Modules $N$ for iterative supervision, and learning rate $\alpha$}
	\KwOutput{Super-resolution image $I_{SR}$}
	Initialize model parameters $\theta_\phi$ and $\theta_\eta$ \\
	\While{training has not converged}{
	    $I_{SR} \leftarrow I_{HR}$ \\
	    \For{n=1 to N}{
    	    Generate LR image $I'_{LR} = \phi(I_{SR}; \theta_\phi)$ \\
	        Evaluate loss  $L_{KANS}(\theta_{\phi}) = ||I'_{LR} - I_{LR}||_p - \beta \| I_{SR} - I_{HR} \|_p + \gamma L_{d}$ \\
    		Update $\theta_\phi \leftarrow \theta_\phi - \alpha \nabla_{\theta_\phi}L_{KANS}(\theta_{\phi})$ 
    		\\ \ \\
    		Generate SR image
    	    $I_{SR} = \eta(I'_{LR}; \theta_\eta)$ \\
    	    Evaluate loss  $L_{SR}(\theta_{\eta}) = \| I_{SR} - I_{HR} \|_p + \omega \| \psi(f(I_{HR})) \odot I_{HR} - \psi(f(I_{SR})) \odot I_{SR} \|_p$ \\
    	    Update $\theta_\eta \leftarrow \theta_\eta - \alpha \nabla_{\theta_\eta}L_{SR}(\theta_{\eta})$
	    }
	}
	
	\caption{The training process of KASR}
	\label{alg}
\end{algorithm}

\section{Experiments}

In the experiments, comparisons between the proposed model with existing SR models are performed. In the ablation study, the effectiveness of each component is validated. Also, we examine the effectiveness of the proposed KASR framework combined with multiple mainstream super-resolution models on two real-world datasets.

\subsection{Experimental Setup}

\noindent \textbf{Evaluation measures.} 
We adopt Peak Signal to Noise Ratio (PSNR)~\citep{yang2014single} and Structural Similarity (SSIM)~\citep{wang2004image} as our evaluation measures. Moreover, since PSNR and SSIM only take pixel-wise distance into consideration, we further include the widely adopted Learned Perceptual Image Patch Similarity (LPIPS)~\citep{zhang2018unreasonable} for perceptual evaluation. LPIPS aims to emulate human observations by depicting the perceptual similarities between SR images and real HR ones.

\noindent\textbf{Datasets.} We conduct model training and evaluation on two real-world image SR datasets, namely: RealSR~\citep{cai2019toward} (including both $\times 2$ and $\times 4$ super-resolution) and City100~\citep{chen2019camera} ($\times 3$ super-resolution). 
In the RealSR dataset, there are 506 and 500 low-resolution/high-resolution image pairs taken by Canon and Nikon cameras for $\times 2$ and $\times 4$ up-scaling, respectively. From RealSR dataset, we adopt 406 images for $\times 2$ model training and 400 images for $\times 4$ model training. The remaining 100 images for each case are useed for model testing. 
The City100 dataset contains 100 real-world low-resolution/high-resolution image pairs taken by a Nikon camera with $\times 3$ up-scaling. We adopt the first 95 images for model training and the remaining 5 images for model evaluation.

\noindent\textbf{Implementation details.} During the training of models, batches of size 32 are adopted, and Adam optimiser is chosen for model optimisation. Following the setup by Lim \etal~\citep{lim2017enhanced}, we randomly flip images vertically/horizontally and rotate them by 90 degrees for data augmentation. The initial learning rate is set to $10^{-4}$. The models are trained for 300 epochs with a multi-step learning rate reduction strategy. The hyper-parameter $\omega$ to compute $L_{SR}(\theta_{\eta})$ in Equation~\eqref{eq:L_SR} is set to 0.5; $\beta$ to compute $L_{KANS}(\theta_{\phi})$ in Equation~\eqref{eq:L_KANS} is set to 1.0; and $\gamma$, also in in Equation~\eqref{eq:L_KANS}, is set to 0.5. For the implementation of the KANS Network, three convolutional layers are combined with the LeakyReLU activation function and Max-pooling operation, where $2 \times 10^{-1}$ is selected as the negative slope of LeakyReLU activation. 32 neural units are set for each hidden layer of the KANS Network. 
In the experiments, the $p$-norm in all losses for the KANS module is defined as the $L_1$ norm, 
and we use $N=2$ KANS modules for the iterative supervision training. 
We use the Sobel filter as the high-frequency filter $f(\cdot)$ and min-max normalisation as $\psi(\cdot)$ in Equation~\eqref{eq:l_hfso}. 
By default, the model labelled as `KASR' in the experiments below consists of our proposed KASR paired with the EDSR model~\citep{lim2017enhanced}. 
The models are implemented with PyTorch and trained/tested on one NVIDIA GTX 1080 Ti graphic card. The experimental settings are kept the same across all experiments for fair comparisons.

\subsection{Overall Performance}

We compare the proposed model with existing super-resolution models. 
Following the mainstream evaluation of SR models~\citep{lim2017enhanced}, we present our model performance with self-ensemble~\citep{timofte2016seven}, which forms the model `KASR+'
, to further enhance the model performance. During the evaluation of `KASR+', instead of inputting a single low-resolution image, we augment it to produce seven images using different rotations, which together with the original image are processed the SR model. After getting the corresponding SR images and rotating the images back to original angles, pixel-level weighted average is performed to obtain the final SR result. The experiments are conducted on RealSR dataset (both $\times 2$ and $\times 4$ up-scalings) and City100 dataset, and results are shown in Tables~\ref{tab:realsrx2-sota},~\ref{tab:realsr-sota}, and~\ref{tab:city100-sota}.

\begin{table}
\centering
\caption{The comparison of our proposed model and existing super-resolution models on RealSR dataset with $\times 2$ up-scaling. The best results per column are bolded, $\uparrow$ indicates that larger is better and $\downarrow$ shows that lower is better.}
\label{tab:realsrx2-sota}
\begin{tabular}{l|c|c|c|c}
\hline
\textbf{Methods} & \textbf{Scale} & \textbf{PSNR $\uparrow$} & \textbf{SSIM $\uparrow$} & \textbf{LPIPS $\downarrow$} \\ \hline
Bicubic & $\times 2$ & 30.270 & 0.874 & 0.210 \\
ZSSR~\citep{shocher2018zero} & $\times 2$ & 30.563 & 0.879 & 0.176 \\
KernelGAN~\citep{bell2019blind} & $\times 2$ & 30.240 & 0.891 & 0.134 \\
MZSR~\citep{soh2020meta} & $\times 2$ & 27.960 & 0.816 & 0.211 \\
DBPI~\citep{kim2020dual} & $\times 2$ & 27.860 & 0.829 & 0.178 \\
DAN~\citep{huang2020unfolding} & $\times 2$ & 30.630 & 0.882 & 0.131 \\
SRResCGAN~\citep{umer2020deep} & $\times 2$ & 26.260 & 0.798 & 0.209 \\
FSSR~\citep{fritsche2019frequency} & $\times 2$ & 30.397 & 0.874 & 0.142 \\
DASR~\citep{wei2021unsupervised} & $\times 2$ & 29.887 & 0.867 & 0.129 \\
CinCGAN~\citep{yuan2018unsupervised} & $\times 2$ & 28.099 & 0.867 & 0.166 \\
ESRGAN~\citep{wang2018esrgan} & $\times 2$ & 30.648 & 0.880 & 0.097 \\ \hline
KASR (Ours) & $\times 2$ & 32.454 & 0.922 & \textbf{0.080} \\
KASR+ (Ours) & $\times 2$ & \textbf{32.519} & \textbf{0.923} & 0.082 \\ \hline
\end{tabular}
\end{table}

\begin{table}
\centering
\caption{The comparison of our proposed model and existing super-resolution models on RealSR dataset with $\times 4$ up-scaling. The best results per column are bolded, $\uparrow$ indicates that larger is better and $\downarrow$ shows that lower is better.}
\label{tab:realsr-sota}
\begin{tabular}{l|c|c|c|c}
\hline
\textbf{Methods} & \textbf{Scale} & \textbf{PSNR $\uparrow$} & \textbf{SSIM $\uparrow$} & \textbf{LPIPS $\downarrow$} \\ \hline
Bicubic & $\times 4$ & 25.740 & 0.741 & 0.467 \\
ZSSR~\citep{shocher2018zero} & $\times 4$ & 26.007 & 0.748 & 0.386 \\
KernelGAN~\citep{bell2019blind} & $\times 4$ & 24.090 & 0.724 & 0.298 \\
DBPI~\citep{kim2020dual} & $\times 4$ & 22.360 & 0.656 & 0.311 \\
DAN~\citep{huang2020unfolding} & $\times 4$ & 26.200 & 0.760 & 0.410 \\
IKC~\citep{gu2019blind} & $\times 4$ & 25.600 & 0.749 & 0.319 \\
SRResCGAN~\citep{umer2020deep} & $\times 4$ & 25.840 & 0.746 & 0.375 \\
FSSR~\citep{fritsche2019frequency} & $\times 4$ & 25.992 & 0.739 & 0.265 \\
DASR~\citep{wei2021unsupervised} & $\times 4$ & 26.782 & 0.782 & 0.228 \\
CinCGAN~\citep{yuan2018unsupervised} & $\times 4$ & 25.094 & 0.746 & 0.405 \\
ESRGAN~\citep{wang2018esrgan} & $\times 4$ & 27.569 & 0.774 & 0.415 \\
RCAN~\citep{zhang2018image} & $\times 4$ & 27.647 & 0.780 & 0.442 \\
Noise-injection~\citep{ji2020real} & $\times 4$ & 25.768 & 0.772 & 0.215 \\ \hline
KASR (Ours) & $\times 4$ & 27.850 & 0.824 & \textbf{0.148} \\
KASR+ (Ours) & $\times 4$ & \textbf{27.991} & \textbf{0.827} & 0.150 \\ \hline
\end{tabular}
\end{table}

We compare our results with the results by multiple strong super-resolution methods obtained from~\citep{chen2022real} and~\citep{wei2021unsupervised} in Tables~\ref{tab:realsrx2-sota} and ~\ref{tab:realsr-sota}. 
The competing methods are:

\begin{itemize}
\item Naive Bicubic up-sampling;
\item ZSSR~\citep{shocher2018zero} and KernelGAN~\citep{bell2019blind} that are degradation estimation models performed on testing data;
\item MZSR~\citep{soh2020meta} that combines Zero-Shot Learning and Meta Transfer Learning;
\item DBPI~\citep{kim2020dual} which is a unified internal learning-based super-resolution model;
\item DAN~\citep{huang2020unfolding} that separates the kernel estimation and restoration as two sub-steps;
\item IKC~\citep{gu2019blind} which iteratively conducts kernel correction; 
\item SRResCGAN~\citep{umer2020deep} that introduces generative adversarial residual CNN into Real-World Super-Resolution domain;
\item FSSR~\citep{fritsche2019frequency} that considers low frequency and high frequency separately for real super-resolution images;
\item DASR~\citep{wei2021unsupervised} which adopts domain-gap aware training strategy and
domain-distance weighted supervision;
\item CinCGAN~\citep{yuan2018unsupervised} that represents Cycle-in-Cycle GAN to use domain translation in real SR domain;
\item ESRGAN~\citep{wang2018esrgan} and RCAN~\citep{zhang2018image} which are two of the most popular models, where ESRGAN is an upgraded version of SRGAN based on the introduction of a residual-in-residual dense block and a new structure, and RCAN proposed a very deep residual network architecture for SR tasks;
\item The Noise-injection model~\citep{ji2020real} that is a baseline real-world super-resolution model. The noise-injection model estimates a finite set of degradation kernels and noises from pre-collected real images by adopting KernelGAN~\citep{bell2019blind} and ZSSR~\citep{shocher2018zero}. In order to deal with real-world SR noises, the noise-injection model performs a degradation operation from the degradation pool during training. However, extra training time for kernel collection is required.
\end{itemize}

On the RealSR results from Tables~\ref{tab:realsrx2-sota} and~\ref{tab:realsr-sota}, our model outperforms competing approaches by a large margin for all three measures, especially on LPIPS evaluations. The DASR model and Noise-injection model have better perceptual results than other competing models, but they are still not as good as our model. On $\times 4$ scaling, compared with the DASR model, the LPIPS obtained by our KASR model is 35\% better.  Compared with the Noise-injection model, our model improves the LPIPS measure from 0.215 to 0.148.

\begin{table}
\centering
\caption{The comparison of our proposed model and existing super-resolution models on City100 dataset with $\times 3$ up-scaling. The best results of a column are bolded, $\uparrow$ indicates that larger is better and $\downarrow$ shows that lower is better.}
\label{tab:city100-sota}
\begin{tabular}{l|c|c|c|c}
\hline
\textbf{Methods} & \textbf{Scale} & \textbf{PSNR $\uparrow$} & \textbf{SSIM $\uparrow$} & \textbf{LPIPS $\downarrow$} \\ \hline
RCAN~\citep{zhang2018image} & $\times 3$ & 28.114 & 0.811 & 0.384 \\
RCAN*~\citep{zhang2018image} & $\times 3$ & 30.016 & 0.864 & 0.260 \\
CinCGAN~\citep{yuan2018unsupervised} & $\times 3$ & 26.221 & 0.703 & 0.303 \\
CamSR-SRGAN~\citep{chen2019camera} & $\times 3$ & 25.257 & 0.764 & 0.195 \\
CamSR-VDSR~\citep{chen2019camera} & $\times 3$ & 30.260 & \textbf{0.868} & 0.263 \\ \hline
KASR (Ours) & $\times 3$ & 30.714 & 0.824 & \textbf{0.112} \\
KASR+ (Ours) & $\times 3$ & \textbf{30.831} & 0.827 & 0.113 \\ \hline
\end{tabular}
\end{table}

Similar results are shown in Table~\ref{tab:city100-sota} on City100 dataset. The competing methods include RCAN and RCAN*~\citep{zhang2018image}, where RCAN* is trained with real low-resolution and high-resolution pairs; CinCGAN~\citep{yuan2018unsupervised}; CamSR-SRGAN~\citep{chen2019camera} and CamSR-VDSR~\citep{chen2019camera} which are two variants of CameraSR model based on SRGAN and VDSR. 
Our model improves the PSNR value from 30.260 to 30.714, compared to the second best model CamSR-VDSR. When compared with CamSR-based models, the proposed model improves the LPIPS measure by a large margin.

Interestingly, from these tables, we can  verify that models with self-ensemble (`KASR+' models) are able to boost performance for both PSNR and SSIM, but not for LPIPS. This can be because the self-ensemble operation performs a pixel-level ensemble that may not effectively increase the perceptual quality of the images. 

\subsection{Ablation Study}

\noindent\textbf{Experiments on RealSR.} We equip the proposed KASR framework on three modern super-resolution models, namely: EDSR~\citep{lim2017enhanced}, SRResNet~\citep{ledig2017photo} and SRGAN~\citep{ledig2017photo}. From Table~\ref{tab:realsr}, we can see that the models equipped with KASR are able to consistently obtain better results, when compared with the original models. Under $\times 2$ up-scaling, SRResNet-KASR increases PSNR by 0.322 compared with the original SRResNet model. SRGAN-KASR also increases the PSNR result by 0.301. For $\times 4$ up-scaling, EDSR-KASR increases the PSNR results of the EDSR model from 27.494 to 27.850 and SSIM from 0.813 to 0.824. Similarly, we notice a PSNR improvement of 0.294 for the SRResNet-KASR model and 0.279 for the SRGAN-KASR model. The performance improvements are not only for the pixel-level evaluation measures (PSNR and SSIM) since the models equipped with KASR are able to produce better results on the perceptual measure LPIPS as well. Empirically, we also notice that SSIM scores are generally more consistent across all models than PSNR results. 
Only on $\times 4$ setting, the SRGAN with KASR model is slightly worse than the baseline in terms of SSIM, but it performs better on the perceptual measure LPIPS, possibly 
because GAN-based models are generally able to capture the semantic information of images than non-GAN models in a better manner.

\begin{table}
\centering
\caption{The comparison between original models and the ones equipped with the proposed KASR framework on RealSR dataset with $\times 2$ and $\times 4$ up-scaling.}
\label{tab:realsr}
\begin{tabular}{l|c|c|c|c}
\hline
\textbf{Methods} & \textbf{Scale} & \textbf{PSNR $\uparrow$} & \textbf{SSIM $\uparrow$} & \textbf{LPIPS $\downarrow$} \\ \hline
EDSR~\citep{lim2017enhanced} & $\times 2$ & 32.378 & 0.922 & 0.081 \\
EDSR-KASR (Ours) & $\times 2$ & 32.454 & 0.922 & 0.080 \\ \hline
SRResNet~\citep{ledig2017photo} & $\times 2$ & 31.630 & 0.915 & 0.086 \\
SRResNet-KASR (Ours) & $\times 2$ & 31.952 & 0.915 & 0.083 \\ \hline
SRGAN~\citep{ledig2017photo} & $\times 2$ & 31.611 & 0.915 & 0.086 \\
SRGAN-KASR (Ours) & $\times 2$ & 31.912 & 0.915 & 0.081 \\ \hline \hline
EDSR~\citep{lim2017enhanced} & $\times 4$ & 27.494 & 0.813 & 0.157 \\
EDSR-KASR (Ours) & $\times 4$ & 27.850 & 0.824 & 0.148 \\ \hline
SRResNet~\citep{ledig2017photo} & $\times 4$ & 27.366 & 0.819 & 0.153 \\
SRResNet-KASR (Ours) & $\times 4$ & 27.660 & 0.820 & 0.153 \\ \hline
SRGAN~\citep{ledig2017photo} & $\times 4$ & 27.355 & 0.819 & 0.156 \\
SRGAN-KASR (Ours) & $\times 4$ & 27.634 & 0.818 & 0.151 \\ \hline
\end{tabular}
\end{table}

\noindent\textbf{Experiments on City100.} Similarly, we equip the KASR framework on EDSR~\citep{lim2017enhanced}, SRResNet~\citep{ledig2017photo} and SRGAN~\citep{ledig2017photo} for the City100 experiment. The quantitative comparisons are shown on Table~\ref{tab:city100}. On the City100 dataset, a 0.743 PSNR increment is obtained by the EDSR-KASR model. We find from this table that the perceptual performance improvements of the model equipped with the KASR framework are clearer on City100 than on RealSR. For instance, on City100, EDSR-KASR improves the LPIPS measure from 0.138 to 0.112; both SRResNet-KASR and SRGAN-KASR improve the LPIPS measure from 0.117 to 0.109. This phenomenon may be caused by the different distribution of data in different datasets.

\begin{table}
\centering
\caption{The comparison between original models and the ones equipped with the proposed KASR framework on City100 dataset with $\times 3$ up-scaling.}
\label{tab:city100}
\begin{tabular}{l|c|c|c|c}
\hline
\textbf{Methods} & \textbf{Scale} & \textbf{PSNR $\uparrow$} & \textbf{SSIM $\uparrow$} & \textbf{LPIPS $\downarrow$} \\ \hline
EDSR~\citep{lim2017enhanced} & $\times 3$ & 29.971 & 0.765 & 0.138 \\
EDSR-KASR (Ours) & $\times 3$ & 30.714 & 0.824 & 0.112 \\ \hline
SRResNet~\citep{ledig2017photo} & $\times 3$ & 30.305 & 0.817 & 0.117 \\
SRResNet-KASR (Ours) & $\times 3$ & 30.309 & 0.816 & 0.109 \\ \hline
SRGAN~\citep{ledig2017photo} & $\times 3$ & 30.181 & 0.815 & 0.117 \\
SRGAN-KASR (Ours) & $\times 3$ & 30.253 & 0.815 & 0.109 \\ \hline
\end{tabular}
\end{table}

\textbf{Effect of different components.} This section examines the contribution of different components of the proposed KASR framework, including the Kernel Adversarial Noise Simulation network (KANS), High-frequency Selective Objective (HFSO) and Iterative Supervision (IS). The quantitative results are shown in the Table~\ref{tab:ablation}.
Starting from the backbone model EDSR in model \#1, we add KANS to build model \#2, which improves PSNR from 27.494 to 27.718; SSIM from 0.813 to 0.823; LPIPS from 0.157 to 0.151. By introducing HFSO and IS, the performance of the model can be further boosted to our best model \#4 with 27.850 on PSNR, 0.824 on SSIM and 0.148 on LPIPS. 
It is worth noting that the KANS module in model \#2 contributes the most to improve the results, producing better results on pixel-level measures (PSNR and SSIM) and the perceptual measure (LPIPS) than model \#1. HFSO and IS also show to further boost model performance.

\begin{table}
\centering
\caption{Ablation study of each proposed component. The experiments are conducted on the RealSR dataset with $\times 4$ up-scaling. The backbone model is EDSR.}
\label{tab:ablation}
\begin{tabular}{c|c|c|c|c|c|c}
\hline
Model \# & KANS & HFSO & IS & PSNR $\uparrow$ & SSIM $\uparrow$ & LPIPS $\downarrow$ \\ \hline
1 & & & & 27.494 & 0.813 & 0.157 \\
2 & \checkmark & & & 27.718 & 0.823 & 0.151 \\
3 & \checkmark & \checkmark & & 27.762 & 0.823 & 0.151 \\
4 & \checkmark & \checkmark & \checkmark & \textbf{27.850} & \textbf{0.824} & \textbf{0.148} \\ \hline
\end{tabular}
\end{table}

\subsection{Analyses}

\noindent\textbf{Number of KANS Modules.} We conduct experiments to examine the number of KANS modules for Iterative Supervision (IS) on Table~\ref{tab:n-modules}. We notice that there is a general trend of evaluation measures improving and then decreasing with a growing number of KANS modules. When number of KANS modules is 2, the performance reaches the peak.

\begin{table}[]
\centering
\caption{The examination of the number of KANS modules for Iterative Supervision (IS). The experiments are conducted on the RealSR dataset with $\times 4$ up-scaling and the backbone model is EDSR. \label{tab:n-modules}}
\begin{tabular}{c|c|c|c}
\hline
KANS modules \# & PSNR $\uparrow$ & SSIM $\uparrow$ & LPIPS $\downarrow$ \\ \hline
1         & 27.762          & 0.823           & 0.151              \\
2         & \textbf{27.850} & \textbf{0.824}  & \textbf{0.148}     \\
3         & 27.764          & 0.823           & \textbf{0.148}              \\ \hline
\end{tabular}
\end{table}

\noindent\textbf{High-frequency Selective Objective Versus High-frequency loss only.} We perform a comparison between the proposed HFSO objective and the loss HF-only computed using only the high-frequency maps of generated SR images and ground-truth images without the element-wise product with the image shown in Equation~\eqref{eq:l_hfso}.
From Table~\ref{tab:HFSO}, the model trained with HFSO shows better results across the three evaluation measures when compared to HF-loss only. In particular,  compared with the HF-only, HFSO increases PSNR from 27.734 to 27.850 and LPIPS from 0.153 to 0.148.
The HFSO objective works more effectively than HF-only since 
it softly weights pixels which enables a richer information content than just using the HF maps.

\begin{table}[]
\centering
\caption{The comparison of the proposed HFSO objective versus high-frequency loss only. The experiments are conducted on the RealSR dataset with $\times 4$ up-scaling and the backbone model is EDSR. \label{tab:HFSO}}
\begin{tabular}{c|c|c|c}
\hline
Objective & PSNR $\uparrow$ & SSIM $\uparrow$ & LPIPS $\downarrow$ \\ \hline
HF-only       & 27.734          & 0.821           & 0.153              \\
HFSO      & \textbf{27.850} & \textbf{0.824}  & \textbf{0.148}     \\\hline
\end{tabular}
\end{table}

\noindent\textbf{Sensitivity of $L_{KANS}(\theta_{\phi})$ to $\beta$ in Equation~\eqref{eq:L_KANS}.} The factor $\beta$ in Equation~\eqref{eq:L_KANS} controls the intensity of adversarial training when computing $L_{KANS}(\theta_{\phi})$, thus playing an important role in the model performance. Table~\ref{tab:sen} shows the model sensitivity to $\beta$.  The experiments are conducted on the RealSR dataset with $\times 4$ up-scaling and backbone model EDSR. It is interesting to note that the results show that our model has a rather stable performance as a function of $\beta$ in terms of PSNR, SSIM and LPIPS measures. 
It is also worth noting that all results show a slight improvement with the increase of $\beta$.
In general, we suggest the use of $\beta=1.0$ to obtain the results shown in this paper.

\begin{table}
\begin{minipage}{.45\linewidth}
\centering
\caption{Sensitivity of $L_{KANS}(\theta_{\phi})$ to $\beta$ in Equation~\eqref{eq:L_KANS} on the RealSR dataset with $\times 4$ up-scaling and backbone model EDSR.}
\label{tab:sen}
\begin{tabular}{c|c|c|c}
\hline
$\beta$ & PSNR $\uparrow$ & SSIM $\uparrow$ & LPIPS $\downarrow$ \\ \hline
$\beta = 0.1$ & 27.722 & 0.823 & 0.150 \\
$\beta = 0.3$ & 27.768 & 0.824 & 0.149 \\
$\beta = 0.5$ & 27.829 & 0.824 & 0.150 \\
$\beta = 0.7$ & 27.795 & 0.824 & 0.150 \\
$\beta = 1.0$ & 27.850 & 0.824 & 0.148 \\ \hline
\end{tabular}
\end{minipage}
\begin{minipage}{.45\linewidth}
\centering
\caption{Sensitivity of $L_{KANS}(\theta_{\phi})$ to $\gamma$ in Equation~\eqref{eq:L_KANS} on the RealSR dataset with $\times 4$ up-scaling and backbone model EDSR.}
\label{tab:sen-gamma}
\begin{tabular}{c|c|c|c}
\hline
$\gamma$ & PSNR $\uparrow$ & SSIM $\uparrow$ & LPIPS $\downarrow$ \\ \hline
$\gamma = 0.1$ & 27.730 & 0.822 & 0.151 \\
$\gamma = 0.3$ & 27.781 & 0.823 & 0.149 \\
$\gamma = 0.5$ & 27.850 & 0.824 & 0.148 \\
$\gamma = 0.7$ & 27.802 & 0.823 & 0.149 \\
$\gamma = 1.0$ & 27.801 & 0.822 & 0.151 \\ \hline
\end{tabular}
\end{minipage}
\end{table}

\noindent\textbf{Sensitivity of $L_{KANS}(\theta_{\phi})$ to $\gamma$ in Equation~\eqref{eq:L_KANS}.} We also examine  the sensitivity of our method to $\gamma$ in Equation~\eqref{eq:L_KANS}, which is the weight of the discriminant loss to classify the original and generated LR images
into real or fake. As shown in Table~\ref{tab:sen-gamma}, the model's performance is quite stable with respect to $\gamma$, with a slight upward trend for $\gamma \in [0.1,0.5]$ followed by a downward trend for $\gamma \in [0.7,1.0]$. Hence,  $\gamma = 0.5$ shows the peak performance, so we recommend $\gamma = 0.5$.

\begin{figure}[]
\centering
\scalebox{1.0}{
\centerline{\includegraphics[width=1\textwidth]{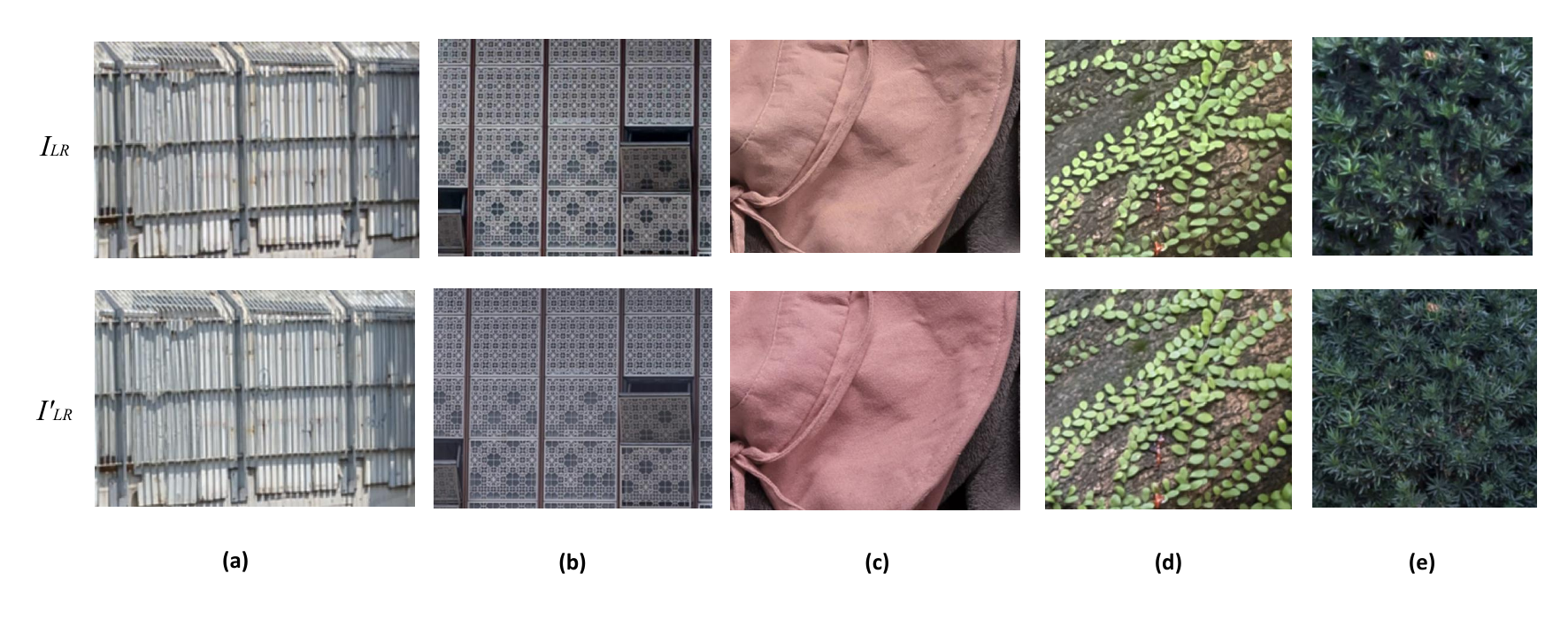}}
}
\caption{Illustration of the differences between original images $I_{LR}$ and KASR produced images $I'_{LR}$.}
\label{fig:gen-img-compare}
\end{figure}

\noindent \textbf{Visualization of generated LR images from KANS.} %\sout{Through sub-figures (a) to (e) in }
Figures~\ref{fig:gen-img-compare}(a)-(e) show comparisons of original low-resolution images $I_{LR}$ and KANS-generated low-resolution images $I'_{LR}$. The distortions include texture changes, colour modifications and blur. As shown in the figure, $I_{LR}$ and $I'_{LR}$ are visually similar, but small perturbations are present in $I'_{LR}$ to mislead deep super-resolution models. 
In Figures~\ref{fig:gen-img-compare}(a), $I'_{LR}$ shows changes in terms of colour and texture when compared with $I_{LR}$. 
In Figures~\ref{fig:gen-img-compare}(b), texture modifications are shown in $I'_{LR}$. 
In Figures~\ref{fig:gen-img-compare}(c), colour changes are displayed on $I'_{LR}$. 
In Figures~\ref{fig:gen-img-compare}(d), blur is shown in $I'_{LR}$ compared with $I_{LR}$. In Figures~\ref{fig:gen-img-compare}(e), some fine details are added to  $I'_{LR}$.

\begin{figure}[]
\centering
\scalebox{1.0}{
\centerline{\includegraphics[width=1\textwidth]{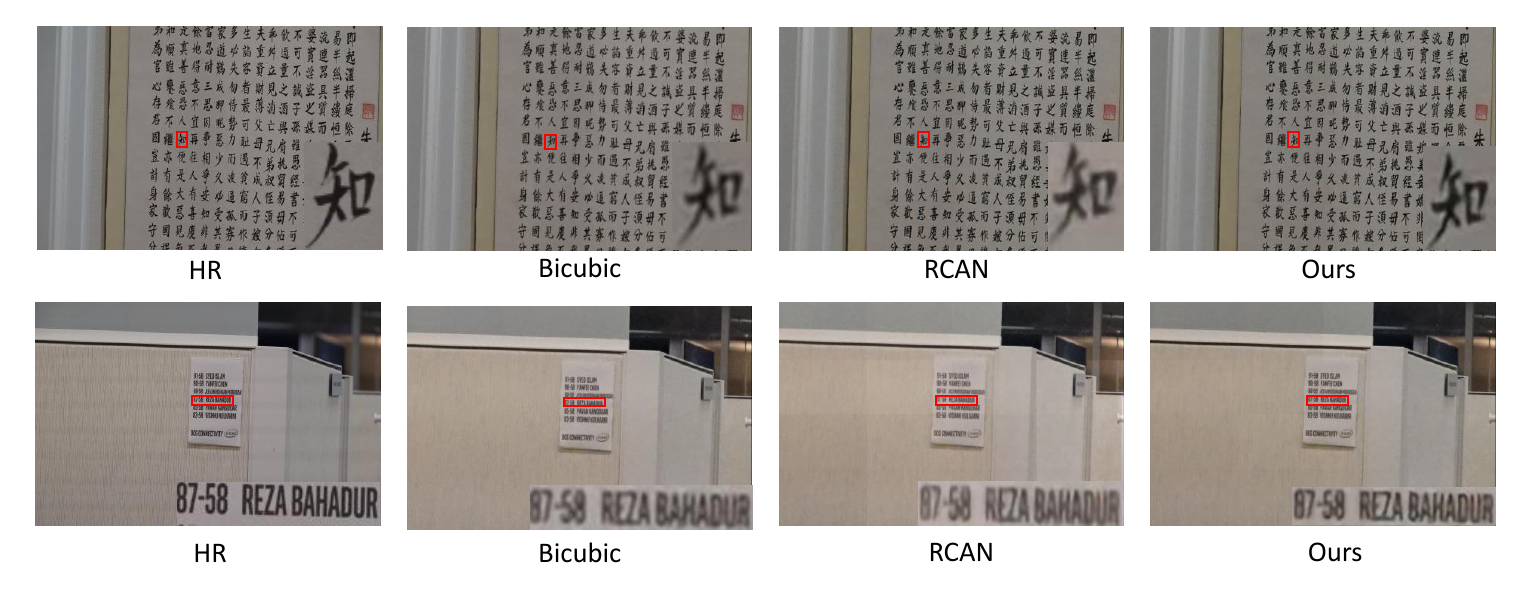}}
}
% \vspace{-2mm}
\caption{Visual comparison between the original HR image, Bicubic interpolation, RCAN and our proposed KASR model.}
\label{fig:sr-vis}
% \vspace{-2mm}
\end{figure}

\begin{figure}[]
\centering
\scalebox{1.0}{
\centerline{\includegraphics[width=1\textwidth]{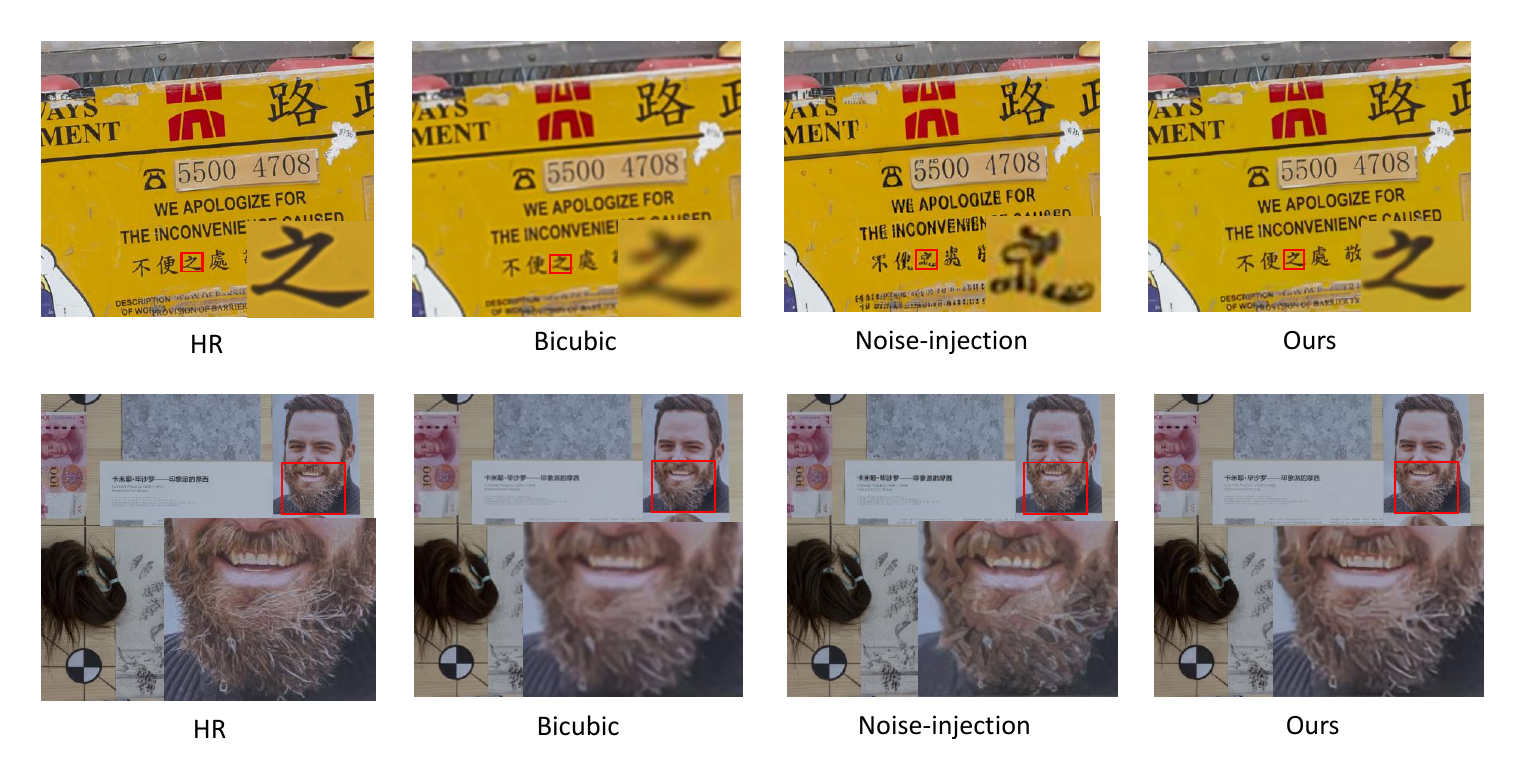}}
}
% \vspace{-2mm}
\caption{Visual comparison between the original HR image, Bicubic interpolation, Noise-injection and the proposed model.}
\label{fig:sr-vis-2}
% \vspace{-2mm}
\end{figure}

\noindent\textbf{Visualization of SR images.} From the visualization in Figure~\ref{fig:sr-vis}, we notice that the proposed model KASR can generate clearer SR images than the Bicubic method and RCAN model in real-world image scenarios. 
Similar results are shown in Figure \ref{fig:sr-vis-2}, where our KASR model shows clearer SR images than the competitors. 
It is worth noting that the Noise-injection model~\citep{ji2020real} can generate clear SR images in some cases, but it can create small artifacts as the character and the man's beard shown in the figures. This may be because its estimated kernels and noises cannot cover all the real degradation kernels.

\section{Conclusion}

In this paper, we proposed the Kernel Adversarial Learning Super-resolution (KASR) to assist SR models to implicitly simulate the complex image degradation process. 
Besides, an iterative supervision process and a high-frequency selective objective were also proposed to further boost the model performance. The experiments combining KASR with three SR models indicated that our proposed framework can be equipped into modern SR models seamlessly. Promising results are shown in real-world SR scenarios by KASR when compared with other models. Furthermore, an extensive ablation study was performed to examine the effectiveness of each component of the proposed framework. From the visualisation of SR images, the proposed approach is able to generate clear SR images compared with its competitors.

For future work, there are several directions that can be explored to further improve the proposed Kernel Adversarial Learning Super-Resolution (KASR) framework. Firstly, investigating advanced Artificial Intelligence-Generated Content (AIGC) techniques. It can further enhance the generation of the degradation kernels and noises as the recent AIGC models show promising results in a wide variety of areas. Besides, exploring the integration of domain-specific knowledge, such as prior information about image degradation processes for specific fields or characteristics of the imaging devices, could help to refine the modelling of real-world degradation. Moreover, extending the framework to handle other image restoration tasks beyond super-resolution, such as denoising or deblurring, would be an interesting direction to explore.

\section*{Acknowledgments}
This project received grant funding from the Australian Government through the Medical Research Future Fund - Public Health Research Development Infrastructure PHRDI 000014 Grant and the Australian Research Council through grants DP180103232 and FT190100525.

\bibliographystyle{plainnat}
\bibliography{mybib}% common bib file
%% if required, the content of .bbl file can be included here once bbl is generated
%%\input sn-article.bbl

\end{document}